\documentclass[conference]{IEEEtran}
\IEEEoverridecommandlockouts
\usepackage{cite}
\usepackage{amsmath,amssymb,amsfonts}
\usepackage{algorithmic}
\usepackage{graphicx}
\usepackage{textcomp}
\usepackage{xcolor}

\usepackage{amssymb}
\usepackage{booktabs}
\usepackage{amsmath}
\usepackage{multirow}
\usepackage{siunitx}
\usepackage{array}
\usepackage{tabularx}
\usepackage{graphicx}
\usepackage{todonotes}
\usepackage{xcolor}
\usepackage{caption}
\usepackage{subcaption}
\usepackage{xspace}
\usepackage{xcolor}
\usepackage{cite}
\newcolumntype{Y}{>{\centering\arraybackslash}X}
\usepackage{wrapfig}

\usepackage[hidelinks]{hyperref}
\usepackage[nameinlink,noabbrev]{cleveref}

\definecolor{ForestGreen}{RGB}{34,139,34}
\definecolor{steelblue}{RGB}{40, 90, 130}
\definecolor{navyblue}{RGB}{0, 0, 128}

\def\BibTeX{{\rm B\kern-.05em{\sc i\kern-.025em b}\kern-.08em
    T\kern-.1667em\lower.7ex\hbox{E}\kern-.125emX}}
\begin{document}

% \title{SD-RouteFusion: End-to-End Ego Trajectory Prediction Using SD Maps}
\title{SD-RouteFusion: Ego-Trajectory Prediction with SD-Map Route Conditioning}

\author{
\IEEEauthorblockN{
Sviatoslav Voloshyn\textsuperscript{*},
Bruno K. W. Martens\textsuperscript{*},
Wangxin Liu,
Jakob Vinkås,
Junsheng Fu
}
\IEEEauthorblockA{
Zenseact, Gothenburg, Sweden \\
\textsuperscript{*}Equal contribution \\
\{sviatoslav.voloshyn, bruno.martens, wangxin.liu, jakob.vinkas, junsheng.fu\}@zenseact.com
}
}

\maketitle

\begin{abstract}
This paper presents {\methodname}, a deployable end‑to‑end ego‑trajectory prediction method that fuses a front-facing camera, vehicle kinematics, and a navigation route derived from a Standard Definition (SD) map. Unlike approaches that rely on High Definition (HD) map geometry, SD‑RouteFusion aligns the learning objective with scalable and production-ready SD-map route inputs, enabling route‑aware prediction without requiring HD‑map infrastructure.
First, we demonstrate that SD‑map route prior provides a powerful long‑horizon semantic prior. Through a comprehensive study on a large‑scale real‑world dataset comprising 480k driving scenarios across 10 European countries and the U.S., we quantify the value of SD‑route conditioning: incorporating SD‑map routes yields a 10.5\% ADE improvement over an image‑and‑kinematics baseline, while our full fusion strategy achieves a 16.9\% ADE reduction given a prediction horizon of 8 seconds. The fusion strategy consists of a dual‑hypothesis design paired with a gated classifier, to ensure robustness under route corruption and visual uncertainty
Finally, to support broader evaluation,  we release an SD‑route generation toolkit (\href{https://github.com/zenseact/SD-RouteFusion}{code link}) that enables SD‑route‑conditioned ego-trajectory prediction on all datasets containing ego pose and future trajectories. Together, {\methodname} establishes a practical path toward robust, route‑aware ego‑trajectory prediction at scale.

%an end-to-end approach for ego-vehicle trajectory prediction that fuses a front-facing camera, vehicle kinematics, and a navigation route derived from Standard Definition (SD) maps. We focus on ego-trajectory prediction (future ego motion) rather than multi-agent motion forecasting, aligning the problem with learned motion planning. Leveraging a large-scale real-world dataset, we show that navigation-aligned SD-route conditioning substantially improves long-horizon ego-trajectory prediction, reducing Average Displacement Error (ADE) by 16.9\% over an image+kinematics baseline. Unlike HD-map-dependent approaches, \methodname uses an SD-map route signal that matches what is available in production vehicles. To remain reliable under route noise (e.g., localization drift or stale maps), we introduce a dual-hypothesis design with late-stage hard gating that selects the more trustworthy prediction at inference time. To facilitate future research, we release a development toolkit and code for SD-map route retrieval from vehicle pose and ground-truth trajectories, enabling evaluation on widely used driving datasets with ego poses and future trajectories.

\end{abstract}

% Junsheng
% Abstract: (1) highlight 580k sceinarioa from fleet data cross EU and US.
% (2) just say baselines methods instead of image+kinematics which is not easy to describe in a clear way for reader to understand.
% (3) Hightlighting we are not using HDmap, which is a common approach in SOTA methods.


\section{Introduction}
Ego-trajectory prediction, which refers to forecasting the future motion of the ego vehicle, is a fundamental problem for autonomous driving, with direct impact on safety and comfort in learned motion planning systems. In this work, we study \emph{ego-trajectory prediction} from onboard sensing and map context. This differs from general motion forecasting, which predicts trajectories for multiple agents (vehicles, pedestrians) given tracked detections. Ego-trajectory prediction is closely related to learned motion planning, where route prior and vehicle dynamics play a central role, and methods typically rely on a combination of onboard sensors and map information \cite{singh2023trajectory, ding2023incorporating}.

Importantly, the goal of this work is not to replace tactical planning or low-level control. Instead, we treat ego-trajectory prediction as the estimation of a likely future ego path conditioned on scene context, vehicle state, and route intent, which can serve as a lightweight prior for downstream planning or safety/intervention systems.

Despite recent progress driven by large real-world datasets \cite{ettinger2021waymo, caesar2020nuscenes, caesar2022nuplan, wilson2023argoverse2, houston2021lyftlevel5}, this task remains challenging due to mismatches between benchmark assumptions and deployable sensing. Many approaches evaluated on forecasting benchmarks consume high-quality object detections and tracks, often generated offline \cite{ettinger2021waymo}, which may not reflect in-vehicle perception failures and missed detections. End-to-end methods that learn directly from raw sensors mitigate this dependency, but are supported by fewer datasets \cite{liu2024survey}. Moreover, because perception errors are handled implicitly in end-to-end pipelines \cite{shi2025motion}, direct comparison to approaches that consume explicit intermediate representations is difficult: forecasting datasets frequently provide tracks and map context without raw camera input, while camera-based driving datasets are typically designed for closed-loop planning and often assume HD-map and multi-sensor inputs. As a result, open-loop ego-forecasting comparisons under a monocular regime are non-trivial. Accordingly, we evaluate ego-trajectory prediction in a deployable sensing setup with a front-facing camera, ego-vehicle state, and a navigation route, and we perform controlled ablations and baseline comparisons on a single consistent dataset.

Existing approaches can be broadly categorized by their reliance on map fidelity. A large body of work leverages lane-level HD maps, which significantly enhance prediction performance \cite{deo2022pgp, schmidt2023navmap, zhou2022hivt}. HD maps provide a strong semantic prior for how a traffic scene will evolve through centimeter-level accurate topology and smooth geometry \cite{liao2024osm}. However, obtaining and maintaining HD maps remains cost- and time-intensive, as online mapping solutions are still insufficient and difficult to scale \cite{schmidt2023navmap, Lilja_2024_CVPR}. Even where coverage is available, road topology changes frequently due to roadworks, resulting in significant remapping costs \cite{luo2023augmenting}. Consequently, limited scalability of HD maps restricts the operational domain of HD-map-reliant methods and widens the gap between academic benchmarks and fleet-scale deployment.

In contrast, production vehicles routinely provide high-level \emph{navigation-route priors} (hereafter referred to as \emph{route priors}) from an onboard navigation system. This signal is lightweight to query and naturally aligned with the ego-vehicle planning stack, yet navigation-route priors are underused in end-to-end ego-trajectory prediction. Navigation routes can be derived from globally available Standard Definition (SD) maps, which are extensively used for online mapping tasks \cite{luo2023augmenting}. SD maps provide road-level rather than lane-level accuracy. They identify major roads but do not distinguish between lanes. In addition, their geometry is less accurate and less smooth than HD maps. Despite these limitations, SD maps are globally available, affordable, and scalable, making them a promising source of semantic context for real-world deployment. Prior evidence suggests SD maps can recover much of the performance gain typically attributed to HD maps over mapless approaches \cite{liao2024osm}. At the same time, navigation and SD-map-derived routes can be imperfect due to localization drift, stale map data (e.g., roadworks), or connectivity errors. This motivates ego-trajectory predictors that can exploit route intent when it is informative, while remaining robust when the route prior is wrong.

To overcome these challenges, we propose \methodname, an end-to-end ego-trajectory predictor that fuses a front-facing camera, ego kinematics, and a navigation-route prior derived from SD maps. The key insight is to treat route intent as a long-horizon prior that can disambiguate future motion, while explicitly accounting for route corruption by producing complementary route-led and image-led trajectory hypotheses. As a result, \methodname can leverage route guidance when it is consistent with the scene and fall back to vision when the route is unreliable. We evaluate the proposed architecture and its ablations on an internal extension of the Zenseact Open Dataset (ZOD) \cite{alibeigi2023zenseact}.

Our contributions are threefold: 
\begin{itemize}
    \item \textbf{Deployable SD-map-conditioned trajectory prediction.}
    We propose \methodname, an end-to-end ego-trajectory prediction framework that conditions on navigation routes derived from globally available SD maps. Aligning the learning objective with scalable and production‑ready SD-map route signals eliminates the need for HD‑map lane geometry and ensures availability.

    \item \textbf{Quantitative evidence of SD-route priors and robust fusion.} We systematically quantify the value of SD-map route intent as a long-horizon semantic prior on a large-scale real-world dataset of 480k driving scenarios collected across 10 European countries and the U.S. Conditioning on SD routes yields a 10.5\% ADE reduction over an image-and-kinematics baseline, while our full fusion strategy achieves a 16.9\% ADE reduction. We design a gating classifier to ensure robustness against route corruption and incorrect route intent.

    \item \textbf{SD-route generation toolkit for broader evaluation.} We release an SD-map route generation toolkit that enables SD-route-conditioned trajectory prediction on other datasets containing ego poses and future trajectories, facilitating reproducible evaluation and future research.\footnote{\url{https://github.com/zenseact/SD-RouteFusion}}
\end{itemize}

\section{Related Works}
\label{sec:relworks}
Ego-trajectory prediction estimates the future motion of the ego vehicle and is closely related to learned motion planning.
In contrast, multi-agent motion forecasting predicts trajectories for surrounding agents, often conditioned on tracked detections and map context. Since our goal is ego-trajectory prediction from onboard sensing and SD map-derived route intent, we focus the remainder of this section on end-to-end and route-conditioned ego-centric approaches.

Learning-based trajectory prediction approaches can be broadly categorized into explicit and implicit modeling approaches \cite{hagedorn2023rethinking}. Explicit methods typically transform input features into interpretable intermediate representations or directly use them to predict future trajectories. In contrast, implicit methods adopt an end-to-end approach, learning a feature space from raw sensor data and/or relevant map information to infer the ego vehicle’s future trajectory.

While explicit methods have been widely used, recent advancements have increasingly favored end-to-end methods \cite{hu2023planning, liang2020pnpnet, chitta2022transfuser, hwang2024emma, gu2023vip3d}, which, despite their potential lack of explainability, have notable advantages. One key benefit is that these methods preserve all available information during the modeling process, avoiding the loss of subtle cues inherent in explicit approaches \cite{hagedorn2023rethinking}. For example, in scenarios where a vehicle is occluded, an end-to-end approach that processes vision data may still detect the vehicle's presence through its shadow on the road. Similarly, the gaze direction of a pedestrian crossing in front of a vehicle can provide important behavioral cues, which may be lost when reducing the pedestrian to a simple bounding box. These capabilities make end-to-end approaches increasingly popular in real-world deployments.

As a result, the focus is shifting from approaches relying on predefined dynamic objects at inference time (e.g., \cite{wilson2023argoverse2}) toward models that reason directly from perception outputs in more realistic, dynamic environments.

\paragraph{End-to-End Trajectory Prediction}
End-to-end methods typically take raw sensor data as input, and usually generate intermediate outputs from a perception module themselves. These inputs often include single-view \cite{hagedorn2023rethinking} or multi-view \cite{hu2023planning, hwang2024emma, gu2023vip3d} camera images, lidar scans \cite{liang2020pnpnet}, or a combination of both modalities \cite{chitta2022transfuser, chen2022learningvehicles}. Additional inputs may include high-level navigation commands (e.g., ``turn left'', ``go straight'') \cite{hwang2024emma} and HD maps \cite{liang2020pnpnet, gu2023vip3d}. Some models also perform temporal fusion over multiple frames, requiring sequences of images rather than single snapshots \cite{gu2023vip3d}.

Despite being end-to-end, many of these models adopt a multitask learning setup, incorporating auxiliary losses such as object detection \cite{hu2023planning, liang2020pnpnet, chitta2022transfuser, chen2022learningvehicles, gu2023vip3d}, online map estimation \cite{hu2023planning, chitta2022transfuser}, occupancy prediction, depth estimation, or semantic segmentation \cite{chitta2022transfuser, hu2023planning}. As a result, these approaches require datasets with extensive ground-truth annotations to supervise each auxiliary task. Although many of such datasets are publicly available, the requirement for rich and diverse supervision can hinder scalability across datasets, especially when transferring to new environments or sensor configurations. For instance, \cite{chitta2022transfuser, chen2022learningvehicles} are trained exclusively on CARLA simulation data, in part because it conveniently provides all the supervision signals required by their multitask setup.

\paragraph{Ego-Centric Trajectory Prediction and Learned Planning}
Several recent works explicitly study ego-trajectory prediction (often framed as open-loop planning) directly from onboard sensing, where the output is the future ego path rather than multi-agent forecasts. For example, Trajectory-guided Control Prediction (TCP) predicts an ego trajectory and uses it to guide control prediction in an end-to-end driving pipeline \cite{wu2022tcp}. Ego Status \cite{li2024egostatus} analyzes how far open-loop end-to-end driving can be pushed using compact ego-state representations and trajectory supervision. Closer to our use of onboard cues, Akbiyik et al.\ leverage driver field-of-view signals for multimodal ego-trajectory prediction \cite{akbiyik2025fov}. In contrast to these works, we focus on fusing camera and kinematics with a \emph{navigation-route prior} derived from globally available SD maps, and explicitly address route corruption via a dual-hypothesis design with late-stage gating. Direct quantitative comparison to these ego-centric works is challenging because they are typically evaluated under different supervision and input assumptions (e.g., privileged driver signals, closed-loop control objectives, or map/label availability), whereas our setting targets open-loop ego-trajectory prediction from front-camera and ego state augmented with an SD-map-derived navigation-route prior. To our knowledge, prior ego-centric end-to-end approaches do not study SD-map-derived navigation routes as an explicit conditioning signal paired with a robustness mechanism that can reject corrupted route priors.

\paragraph{Map Representations for Trajectory Prediction}

HD maps are popular as a cue for motion prediction \cite{deo2022pgp, zhou2022hivt, hagedorn2023rethinking}, and have been shown to significantly increase prediction performance \cite{hagedorn2023rethinking}. This improvement currently comes at the cost of the operational design domain of these methods, given the scalability challenges posed by HD maps \cite{luo2023augmenting}. 
A little-studied alternative is using SD maps for motion prediction tasks. One explicit method \cite{schmidt2023navmap} shows that it significantly increases prediction performance compared to a map-free approach and approximates performance of methods leveraging HD maps. However, \cite{schmidt2023navmap} is not fully HD map agnostic as the training phase of their learning-based approach leverages knowledge distillation in a teacher-student model setup, where the teacher relies on HD maps. This limits the diversity of the training data. Another study \cite{liao2024osm} demonstrates that using SD maps as opposed to HD maps does not lead to a significant decrease in performance in a state-of-the-art prediction method \cite{zhou2022hivt}. This motivates further research in this direction, given the significant scalability benefits it provides. However, contrary to \methodname, the aforementioned approaches do not include route-conditioning and are not trained end-to-end.

\paragraph{Conditioning Predictions} Another emerging trend in trajectory prediction is conditioning predictions based on a goal or path \cite{hagedorn2023rethinking, gu2021densetnt}. Unlike TNT/DenseTNT-style goal-conditioned forecasting \cite{zhao2021tnt, gu2021densetnt}, which primarily targets multi-agent prediction with candidate goal hypotheses, our setting is ego-centric and leverages a navigation-route prior (path intent) available in production vehicles. Several methods have improved prediction performance by using a goal or path as a semantic prior for the future path/trajectory of a vehicle \cite{ding2023incorporating}. Many methods predict these goal locations \cite{gilles2022gohome, zhao2021tnt, gu2021densetnt}. An alternative is relying on navigation information, such as presented in \cite{caesar2022nuplan, afsharpbp}. Although, goal conditioning has proven to be very effective \cite{zhao2021tnt, afsharpbp}, recently route/path conditioning has gained popularity, as it not only accounts for the feasibility of the goal position, but also the path to be traversed to reach that goal location \cite{deo2022pgp}. Moreover, a method that only encodes the part of the map that corresponds with the intended route reached promising results \cite{hallgarten2023prediction}. This approach could be effectively combined with using SD map-based navigation, which can be directly retrieved from the on-board navigation system in a real-world setting, making it an interesting and valuable direction for future research.

\section{Methodology}
\label{sec:methodology}
In this section, we describe our methodology. The route generation algorithm is described in \autoref{sec:route_generation}, while the rest of the model architecture is discussed in \autoref{sec:model_architecture}.

\begin{figure*}[t]
\centering
\fcolorbox{lightgray}{white}{%
    \includegraphics[width=\textwidth]{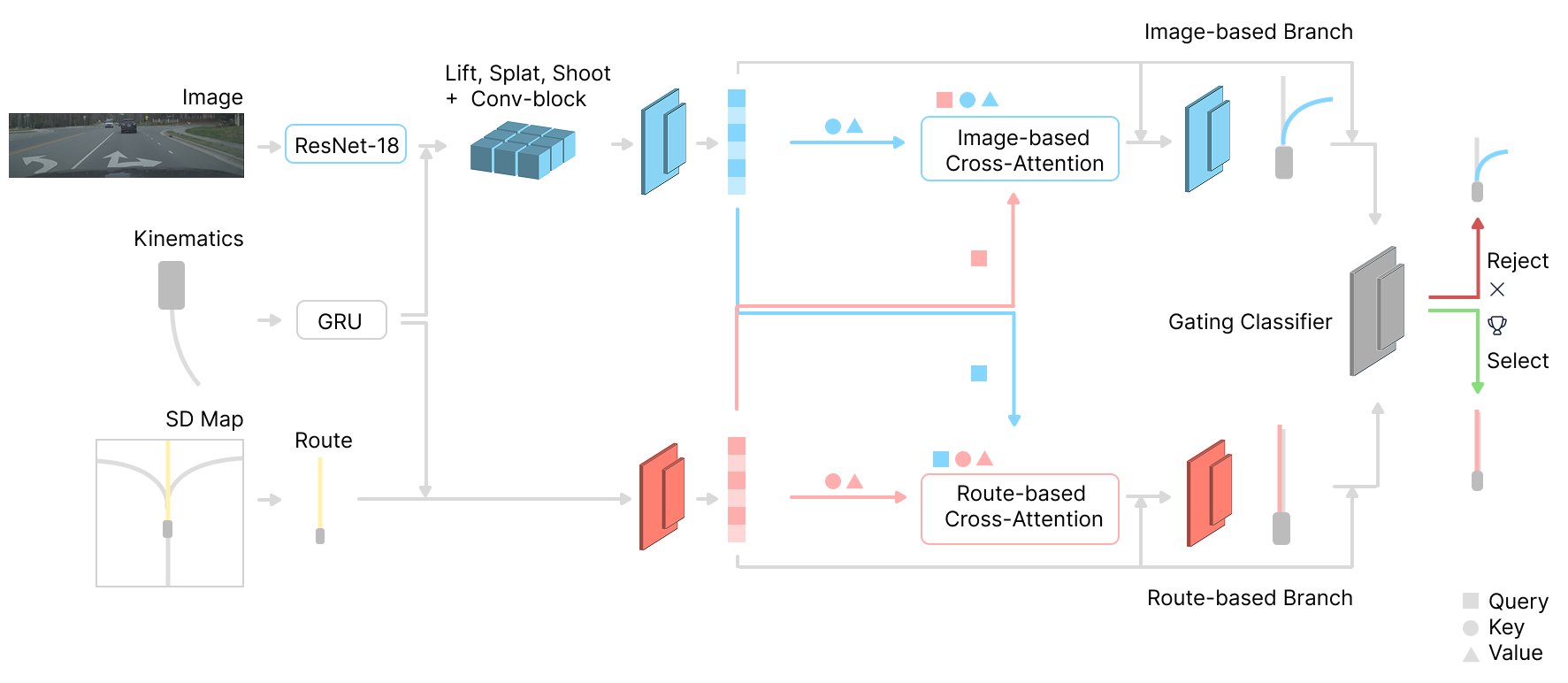}%
}
\caption{Schematic overview of \methodname. Inputs are the navigation route retrieved from SD map, the front-facing camera image and ego-vehicle kinematics. Trajectory predictions are performed in two branches, either image- or route-based, and a gating classifier selects the final prediction. The cross-attention blocks allow for information fusion from the leading modalities to the other branch. Lift, Splat, Shoot \cite{philion2020lift} is used to lift image features to a birds-eye-view perspective to align with the other modalities.}
\label{fig:main}
\end{figure*}

\subsection{SD-Map Route Generation}
\label{sec:route_generation}

To simulate in-vehicle navigation, we construct route information based on the start and end points of the ground truth trajectory, obtained from a high-precision GNSS device (OxTS) and projected onto the nearest SD map elements. Given the initial position $p_{\text{init}}$ and final position $p_{\text{final}}$, we query OpenStreetMap \cite{Vargas_Munoz_2021} for all SD map links within a custom radius of \SI{20}{\meter}, denoting the resulting set as $L$. If no links are found, we incrementally expand the radius in steps of \SI{50}{\meter} up to a maximum of \SI{1}{\kilo\meter}. If no valid links are retrieved within this range, we treat it as a "no-route" case and return a straight-line fallback route. This typically occurs in case of extreme localization issues, node connectivity errors within a map or simply outdated information. In all other cases, we define the reference path vector as $P_{\text{ref}} = p_{\text{final}} - p_{\text{init}}$ and compute the route as follows.

For each map link $l_i \in L$, we determine the point $p_i$ on $l_i$ closest to $p_{\text{init}}$, forming the set $Q = \{p_i\}_{i=1}^{|L|}$. Starting from each point $p_i \in Q$, we simulate all possible traversals across the map, respecting topology constraints such as connectivity and driving direction, thereby generating a set of candidate routes $O = \{o_i\}$.
Each candidate route $o_i$ consists of a sequence of points $\{o_{i,j}\}_{j=1}^{n_i}$, where $n_i$ is the number of points along the route $o_i$.

We select the final route $R$ by minimizing the average displacement error (ADE) between each candidate route $o_i$ and the reference vector $P_{\text{ref}}$, evaluated only up to the projection of $p_{\text{final}}$ onto $o_i$:
\begin{equation}
\text{ADE}(o_i) = \frac{1}{n_i}\sum_{j=1}^{n_i} \left\| o_{i,j} - \left(p_{\text{init}} + \frac{j}{n_i} P_{\text{ref}} \right) \right\|_2,
\end{equation}

\begin{equation}
R = \arg\min_{o_i \in O} \text{ADE}(o_i),
\end{equation}
where $\|\cdot\|_2$ denotes the Euclidean distance.
The number of points $n_i$ is small enough to ensure sufficient spatial resolution along each candidate route. After selecting the best-matching route, we re-center it relative to the vehicle's starting position by subtracting the first route point, ensuring that the route begins at the origin $(0, 0)$. This is done to minimize the influence of any localization errors. In deployment, the same route prior can be obtained directly from an onboard navigation system given the vehicle's current pose and destination; our SD-map procedure serves to emulate this signal consistently in logged datasets.

\subsection{Model Architecture}
\label{sec:model_architecture}
\paragraph{Overview}
As illustrated in \autoref{fig:main}, the proposed architecture consists of two prediction branches and a gating module, which together produce a single final predicted ego trajectory. This output should be interpreted as a route-aware ego-motion prior rather than a complete driving policy or maneuver-planning decision. The model takes as input a forward-facing camera image $I$ and a sequence of kinematic attributes $K \in \mathbb{R}^{n \times 6}$, which encode the 2D position, heading, longitudinal velocity, longitudinal acceleration, and yaw rate over $n$ timestamps. These are aggregated using a gated recurrent unit (GRU) into a compact feature vector $F_k$.

The image $I$ is processed by a truncated ResNet-18 backbone, followed by a bird's eye view (BEV) projection module inspired by Lift-Splat-Shoot \cite{philion2020lift}, and a lightweight convolutional refinement block. The resulting image features are fused with $F_k$ via a multi-layer perceptron (MLP), producing an uncertainty-aware image feature embedding $E_i \in \mathbb{R}^{64}$ that encodes plausible future motion conditioned on the visual scene and vehicle dynamics. In parallel, a route prior $R$ extracted from an SD map \autoref{sec:route_generation} is fused with $F_k$ via a lightweight MLP, yielding a route-conditioned embedding $E_r \in \mathbb{R}^{64}$. This fusion smooths sharp turns and introduces uncertainty when route and kinematic cues diverge.

To resolve ambiguities in the image-based trajectory distribution, we apply a cross-attention mechanism using $E_r$ as queries and $E_i$ as keys and values. This allows the image-based branch to follow global route guidance while preserving fine-grained visual cues. In parallel, a route-based prediction is generated by reversing the attention direction by using $E_i$ as queries and $E_r$ as keys and values. This prioritizes high-level route consistency while still allowing local scene-driven adjustments.

Finally, a gating module takes as input $F_k$, $E_i$, and $E_r$, and the disagreement between the two trajectory predictions. It learns to dynamically select the more reliable branch at inference time, rather than merging both, improving robustness in cases of conflicting inputs.

\paragraph{Cross-Attention}
\label{sec:cross_attention}

We use mirrored cross-attention to exchange information between route and image embeddings. Each operation designates one modality as \emph{query} and the other as \emph{key/value}. Specifically, in the route-updated branch we compute $\mathrm{Attn}(E_r, E_i)$, where the route embedding $E_r$ provides queries and the image embedding $E_i$ provides keys/values; this injects local visual cues into a route-consistent hypothesis. Conversely, $\mathrm{Attn}(E_i, E_r)$ updates the image-led hypothesis using route intent.
The attention mechanism is formulated as:
\begin{equation}
\text{Attn}(\mathbf{A}, \mathbf{B}) = \text{Softmax}\left(\frac{\text{LN}(\mathbf{A}) \cdot \text{LN}(\mathbf{B})^T}{\sqrt{d}}\right) \cdot \text{LN}(\mathbf{B}),
\end{equation}
where \( \mathbf{A} \) provides queries and \( \mathbf{B} \) provides keys/values. The output of the attention mechanism is then processed through an MLP to generate route-based output $O_r$ and image-based output $O_i$:
\begin{equation}
\begin{aligned}
O_{r} &= \text{MLP}(\text{Attn}(E_r, E_i)),\\
O_{i} &= \text{MLP}(\text{Attn}(E_i, E_r)).
\end{aligned}
\end{equation}
For further refinement, a skip connection from the dominant modality is applied to the output:
\begin{equation}
\begin{aligned}
T_{\text{r}} &= \text{MLP}(O_{\text{r}} + E_r),\\
T_{\text{i}} &= \text{MLP}(O_{\text{i}} + E_i)
\end{aligned}
\end{equation}
Together, the two attention directions produce route-led and image-led hypotheses that are each informed by the other modality.

\paragraph{Gating}
The motivation for introducing a hard gating module stems from the limitations of soft fusion approaches, which tend to blend modalities rather than explicitly selecting between them. Since both inputs to the gating module are already informed by both image and route contexts, the gating's primary role is to select the more reliable output rather than to merge information. Each branch benefits from the high-confidence visual cues close to the vehicle and the long-range geometric consistency provided by the route. However, the route-based output remains heavily dependent on accurate localization and map information, and can fail catastrophically when these signals are misaligned. In contrast, the image-based output serves as a fallback option, offering greater robustness in the immediate vicinity of the vehicle at the cost of reduced long-range precision. It is the role of gating to select the appropriate output. 

Architecturally, the gating module is a simple MLP block, which takes in embeddings $E_r$ and $E_i$ as well as the difference between branch predictions $T_{r}-T_{i}$.
\begin{equation} 
g = \text{MLP}\left(\left[E_r, E_i, T_{r} - T_{i}\right]\right),
\end{equation}
where $g$ is a logit scalar. If $g>0$, the image-based prediction $T_i$ is selected, otherwise the route-based prediction $T_r$ is used.

\paragraph{Losses}
\label{sec:losses}

Each trajectory prediction branch is supervised individually using an L2 loss against the ground truth trajectory:
\begin{equation}
\begin{aligned}
\mathcal{L}_{\text{image}} = \|T_i - T_{\text{GT}}\|_2,\\
\mathcal{L}_{\text{route}} = \|T_r - T_{\text{GT}}\|_2,
\end{aligned}
\end{equation}
where \( T_i \) and \( T_r \) denote the image-based and route-based trajectory predictions, respectively, and \( T_{\text{GT}} \) is the ground truth trajectory.

The gating module is trained with a self-supervised binary cross-entropy (BCE) loss. We derive the supervision signal by comparing the per-sample L2 errors of the two branches: if the image-based prediction is closer to the ground truth than the route-based one, the gating target favors the image branch, and vice versa. Specifically, we define a soft supervision target:
\begin{equation}
s = \sigma\left(\tau (e_r - e_i)\right),
\end{equation}
where \( e_i = \|T_i - T_{\text{GT}}\|_2 \), \( e_r = \|T_r - T_{\text{GT}}\|_2 \), \( \sigma(\cdot) \) is the sigmoid function, and \( \tau \) is a temperature hyperparameter controlling the sharpness of the target. The final gating loss is then:
\begin{equation}
\mathcal{L}_{\text{gating}} = -(s \log \sigma(g) + (1 - s) \log (1 - \sigma(g))),
\end{equation}
where \( g \) is the predicted gating logit. The total loss combines these objectives as:
\begin{equation}
\mathcal{L} = \lambda_{\text{traj}} \frac{\mathcal{L}_{\text{image}} + \mathcal{L}_{\text{route}}}{2} + \lambda_{\text{gating}} \mathcal{L}_{\text{gating}},
\end{equation}
where \( \lambda_{\text{traj}} \) and \( \lambda_{\text{gating}} \) are scalar weights balancing the trajectory and gating losses.

\section{Experiments}
In this section, we present our experimental setup, compare \methodname's performance to baselines, and conduct ablation studies to assess the contribution of different components.

\begin{figure}[t]
\centering
\begin{minipage}[c]{0.5\columnwidth}
  \centering
  \includegraphics[height=4.1cm,keepaspectratio]{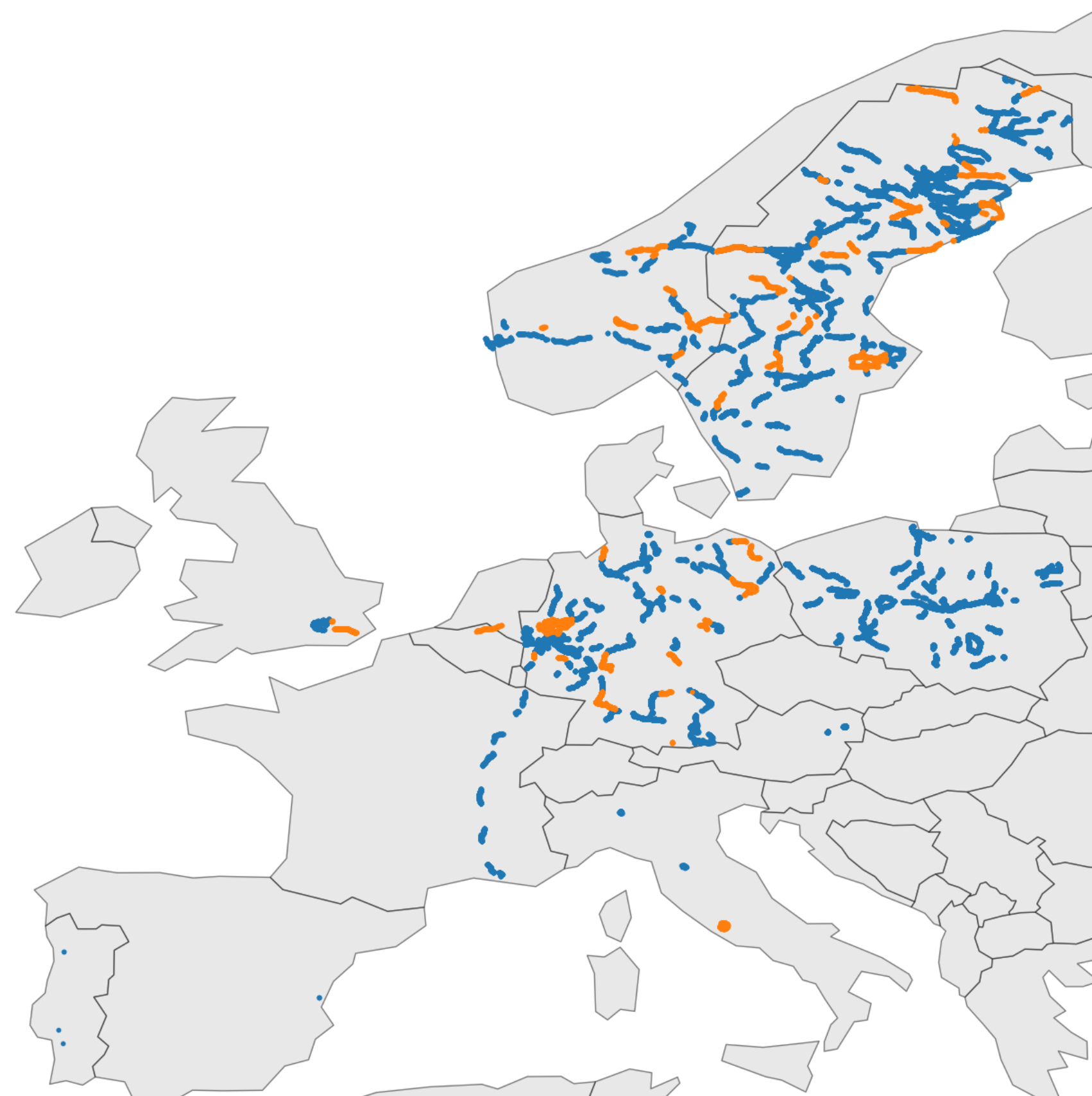}
\end{minipage}\hfill
\begin{minipage}[c]{0.5\columnwidth}
  \centering
  \includegraphics[height=4.1cm,keepaspectratio]{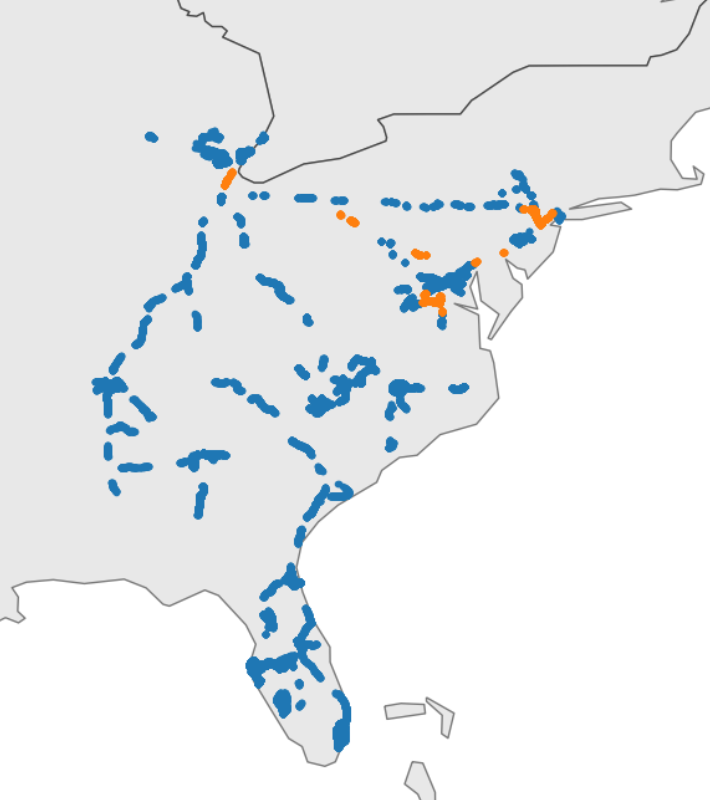}
\end{minipage}
\caption{Geographical coverage of our dataset, split per \textcolor{steelblue}{train} and \textcolor{orange}{test} set.}
\label{fig:dataset-coverage}
\end{figure}

\paragraph{Dataset}
We utilize an extended version of ZOD \cite{alibeigi2023zenseact} consisting of {\totalsamples} scenarios, collected primarily across ten countries in Northern and Central Europe as well as along the East Coast of the United States. The test subset consists of {\evalsamples} samples. The sensor suite includes a forward-facing camera image, along with vehicle position, heading, velocity, and acceleration measurements obtained from an IMU. Each scenario contains at least \SI{1.5}{\second} of ego vehicle kinematic history and ground-truth ego trajectory for the prediction horizon of \SI{8}{\second}, similar to \cite{ettinger2021waymo, caesar2022nuplan}.

\paragraph{Data availability and reproducibility}
Our experiments use an internal extension of the Zenseact Open Dataset (ZOD) \cite{alibeigi2023zenseact} containing {\totalsamples} scenarios. Due to licensing and privacy constraints, this full extension is \emph{not} publicly released. The publicly available ZOD release provides the raw sensor data and vehicle state needed to construct the ego-trajectory prediction setup used in this paper. To support reproducibility, we release (i) the full training and evaluation code for \methodname, and (ii) an SD-map route generation toolkit that, given ego poses and a ground-truth ego trajectory (as provided by ZOD and other datasets), retrieves a local OpenStreetMap extract and computes a navigation-route prior. This enables reproducing the full pipeline and evaluations on the public ZOD split (and porting the approach to other public driving datasets with ego poses and future trajectories), while our reported numbers on the internal extension should be interpreted as results on additional proprietary data.

% --- Dataset comparison table (wider last column, clearer ZOD row) ---
\begin{table*}[t]
\centering
\small
\setlength{\tabcolsep}{3pt}
\renewcommand{\arraystretch}{1.08}

\caption{Comparison of datasets against our required regime (raw \textbf{front camera} + \textbf{$\geq$8\,s} continuous ego future GT for \textbf{open-loop} forecasting). 
For time-based releases, we report an approximate count of non-overlapping 8\,s windows (\(\approx\) hours\(\times 3600/8\)).}
\label{tab:dataset_comparison}

\begin{tabularx}{\textwidth}{lccclX}
\toprule
\textbf{Dataset} & \textbf{Cam} & \textbf{8s} & \textbf{Scale (as released)} & \textbf{Focus} & \textbf{Why not a drop-in benchmark} \\
\midrule

Argoverse 2 MF \cite{wilson2023argoverse2}
& $\times$ & $\checkmark$ & 0 (no images) & forecasting
& No raw camera imagery \(\Rightarrow\) cannot evaluate camera-conditioned ego prediction. \\

Waymo Open Motion \cite{ettinger2021waymo}
& $\sim$ & $\checkmark$ & 0 (no raw video) & forecasting
& Does not release raw camera frames over the prediction window (tokenized image prefix only). \\

nuPlan \cite{caesar2022nuplan}
& $\checkmark$ & $\checkmark$ & \(\approx\)54k--58k (120--128\,h sensor) & planning
& Public raw sensor share is \(\sim\)10\% of logs and scenario-mined for planning; ecosystem is HD-map/closed-loop centric, so dense \emph{monocular} open-loop benchmarking is less standardized and effectively smaller. \\

Waymo End-to-End Driving \cite{xu2025wode2e}
& $\checkmark$ & $\checkmark$ & \(\approx\)4k--5k (segments) & end-to-end
& Closest public match (camera + long horizon), but much smaller. \\

Extended ZOD (ours) \cite{alibeigi2023zenseact}
& $\checkmark$ & $\checkmark$ & \textbf{480k scenarios} & driving logs
& Matches our input regime at scale (front cam + ego state + 8\,s GT). Public ZOD + our SD-route toolkit enable reproduction on the public split. \\

\bottomrule
\end{tabularx}

\vspace{2pt}
\footnotesize{$\checkmark$ available;\; $\times$ not available;\; $\sim$ partial (not raw frames / limited temporal coverage).}
\end{table*}

\paragraph{Benchmarking note}
We compare against strong in-setting baselines (I+K, I+K+R, CVM) because many public benchmarks are not drop-in compatible with our regime (front camera + ego state + 8\,s open-loop ego GT). 
Table~\ref{tab:dataset_comparison} summarizes the key mismatches; our SD-route toolkit enables transferring SD-route conditioning and gating to any dataset with ego poses and future trajectories.

\paragraph{Metrics}

We evaluate performance using three standard metrics: Average Displacement Error (ADE), Final Displacement Error (FDE), and Miss Rate (MR). ADE computes the mean $\ell_2$ distance between predicted and ground truth points across the entire prediction horizon, while FDE considers only the final predicted point. MR captures the proportion of scenarios in which the final prediction lies beyond a fixed threshold from the ground truth. We use a threshold of \SI{4}{\meter}.

\paragraph{Baselines and Ablations} 
To illustrate dataset complexity and contextualize the performance of \methodname, we employ a non-learning-based baseline. Specifically, we evaluate a route-constrained Constant Velocity Model (CVM) \cite{scholler2020constant}. 
The model assumes that the velocity at prediction time remains constant throughout the prediction horizon, but constrains the resulting future motion to follow the retrieved SD-map route rather than extrapolating freely in Euclidean space. While simple, this intuitive baseline sheds light on the complexity of our dataset and helps put our results into context.

We denote image, kinematics, and route inputs as I, K, and R, respectively. Thus, I+K refers to the image-and-kinematics baseline, while I+K+R refers to the early-fusion baseline that additionally receives the SD-map route prior.

Additionally, we perform an ablation study to compare the influence of inputs and architectural choices. First, we fuse our image backbone, ResNet-18, with kinematic features passed through a GRU and decode the trajectory using convolutional blocks and an MLP (I+K). This model shows the strength of our image modality in combination with kinematics, but with no input of route information. Next, as a competitive baseline for \methodname, we add route as an additional input, but decode the trajectory using an MLP, thereby not using the cross-attention blocks and gating classifier as presented in \methodname. Lastly, we also measure the isolated contribution of the gating classifier, by evaluating both prediction branches on the full test set. 

\begin{table*}[t]
\centering
\small
\caption{Prediction performance of models across three scenario types. Prediction horizon is indicated between brackets. MR threshold is \SI{4}{\meter}. I stands for image, K for kinematics and R for route information. Values in brackets denote error reductions relative to the row above.}
\label{tab:experiments-improvements}
\begin{tabularx}{\textwidth}{lYYYYYYYYY}
\toprule
\textbf{Model} 
& \multicolumn{3}{c}{\textbf{Full Test Set (8s)}} 
& \multicolumn{3}{c}{\textbf{Turning Cases (8s)}} 
& \multicolumn{3}{c}{\textbf{Full Test Set (5s)}} \\
\cmidrule(lr){2-4} \cmidrule(lr){5-7} \cmidrule(lr){8-10}
& 
\scalebox{0.9}{\textbf{ADE} $\downarrow$}
 & \scalebox{0.9}{\textbf{FDE} $\downarrow$} & \scalebox{0.9}{\textbf{MR} $\downarrow$}
& \scalebox{0.9}{\textbf{ADE} $\downarrow$}
 & \scalebox{0.9}{\textbf{FDE} $\downarrow$} & \scalebox{0.9}{\textbf{MR} $\downarrow$}
& \scalebox{0.9}{\textbf{ADE} $\downarrow$}
 & \scalebox{0.9}{\textbf{FDE} $\downarrow$} & \scalebox{0.9}{\textbf{MR} $\downarrow$} \\
\midrule
Route-CVM          & 5.10 & 12.02 & 0.80 & 7.48 & 18.04 & 0.92 & 2.43 & 5.20 & 0.62 \\
I + K        & 2.19 & 5.58  & 0.67 & 2.47 & 6.43  & 0.75 & 0.93 & 2.10 & 0.27 \\
I + K + R    & 1.96 & 4.51  & \textbf{0.55} 
             & 2.04 & 4.63  & 0.59 
             & 0.99 & 1.96  & 0.23 \\
             & \multicolumn{1}{c}{\scriptsize(10.5\%)} & \multicolumn{1}{c}{\scriptsize(19.2\%)} & \multicolumn{1}{c}{\scriptsize(17.9\%)} 
             & \multicolumn{1}{c}{\scriptsize(17.4\%)} & \multicolumn{1}{c}{\scriptsize(28.0\%)} & \multicolumn{1}{c}{\scriptsize(21.3\%)} 
             & \multicolumn{1}{c}{\scriptsize(6.5\%)} & \multicolumn{1}{c}{\scriptsize(6.7\%)} & \multicolumn{1}{c}{\scriptsize(14.8\%)} \\
\methodname  & \textbf{1.82} & \textbf{4.32} & \textbf{0.55} 
             & \textbf{1.92} & \textbf{4.50} & \textbf{0.58} 
             & \textbf{0.88} & \textbf{1.85} & \textbf{0.22} \\
             & \multicolumn{1}{c}{\scriptsize(7.1\%)} & \multicolumn{1}{c}{\scriptsize(4.2\%)} & \multicolumn{1}{c}{\scriptsize(0.0\%)} 
             & \multicolumn{1}{c}{\scriptsize(5.9\%)} & \multicolumn{1}{c}{\scriptsize(2.8\%)} & \multicolumn{1}{c}{\scriptsize(1.7\%)} 
             & \multicolumn{1}{c}{\scriptsize(11.1\%)} & \multicolumn{1}{c}{\scriptsize(5.6\%)} & \multicolumn{1}{c}{\scriptsize(4.3\%)} \\
\bottomrule
\end{tabularx}
\end{table*}

\subsection{Quantitative Results}
% We study the performance of our presented models under three scenario subsets. First, we evaluate the performance of the full test dataset. We also evaluate two subsets: turning cases, as measured by a lateral displacement of the ground truth of at least \SI{75}{\meter} and scenarios with a shorter prediction horizon of \SI{5}{\second}, to study the performance under turning behaviour and see the influence of the modalities under varying prediction horizons. The results are shown in \autoref{tab:experiments-improvements}.

We evaluate all models on the full test set, turning cases (ground-truth lateral displacement above \SI{75}{\meter}), and a shorter \SI{5}{\second} prediction horizon to study turning behavior and modality influence across horizons. The results are shown in \autoref{tab:experiments-improvements}.

The numerical results highlight two key findings. First, we observe that the route retrieved from SD maps serves as a strong prior, significantly enhancing performance particularly at longer prediction horizons when compared to a strong image-only baseline. For an \SI{8}{\second} horizon, incorporating the route even in a trivial manner via an MLP leads to a 10.5\% reduction in ADE, and a 19.2\% reduction in FDE. The benefits are more pronounced on the turning-cases subset, where FDE improves by 28.0\%. This gain is likely due to the route's ability to resolve ambiguities caused by occlusions or limited visual context in curved or branching paths.

Second, we find that \methodname is an effective mechanism for integrating route information with image features in a structured and interpretable manner. Using \methodname's cross-attention and gating fusion between the route and image modalities rather than fusing via a simple MLP further improves performance by 7.1\% for ADE and 4.2\% for FDE on the full test set, supporting the design choice.

\begin{table}[t]
\centering
\small
\caption{Ablation study on the full test set with a prediction horizon of \SI{8}{\second}. MR threshold is \SI{4}{\meter}.}
\label{tab:experiments2}
\begin{tabular}{lccc}
\toprule
\textbf{Model} 
& \textbf{ADE} $\downarrow$ 
& \textbf{FDE} $\downarrow$ 
& \textbf{MR} $\downarrow$ \\
\midrule
Image-based Branch & 2.59 & 6.59 & 0.73 \\
Route-based Branch & 1.91 & 4.47 & \textbf{0.55} \\
\methodname        & \textbf{1.82} & \textbf{4.32} & \textbf{0.55} \\
\bottomrule
\end{tabular}
\end{table}

These results suggest the task benefits from selecting between distinct trajectory hypotheses rather than blending them, particularly under route corruption where early fusion can over-trust a misaligned prior, supporting our use of late-stage gating. Empirically, this approach performs well, outperforming early fusion baselines. It combines strengths of both modalities and achieves route-level accuracy at long horizons while maintaining image-level reactivity in the short term.

Route priors derived from SD maps are noisy in real deployments due to localization drift and map inconsistencies (e.g., roadworks), which can misalign the route with the true drivable path. \methodname explicitly addresses this by generating route-led and image-led hypotheses and using late-stage gating to select the more reliable one at inference time. The gating chooses the route-led branch in 66\% of test scenarios, but can fall back to vision when the route prior is corrupted, as illustrated by the mislocalization and outdated-map examples in \autoref{modelscomp}. The partial-occlusion cases in \autoref{fig:partial_predictions} further show that route guidance resolves visual ambiguity, while failures typically occur only when both modalities are simultaneously misleading (\autoref{fig:partials_failure}).

\subsection{Qualitative Results}

\begin{figure*}[t]
\centering
\begin{subfigure}[b]{0.32\textwidth}
    \includegraphics[width=\linewidth]{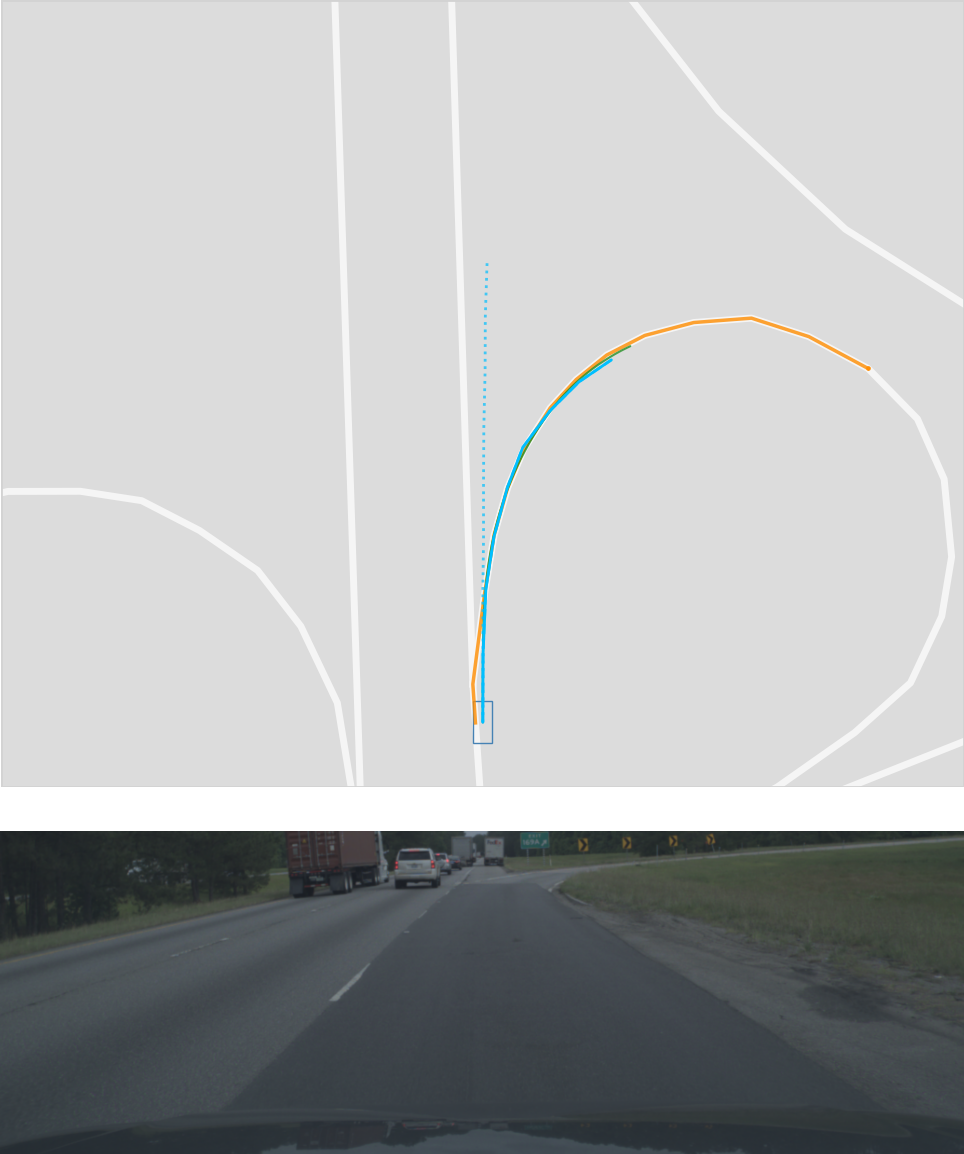}
    \caption{Gating correctly resolves ambiguity in the image, prioritizing route-based prediction.}
    \label{fig:partials_good_1}
\end{subfigure}\hfill
\begin{subfigure}[b]{0.32\textwidth}
    \includegraphics[width=\linewidth]{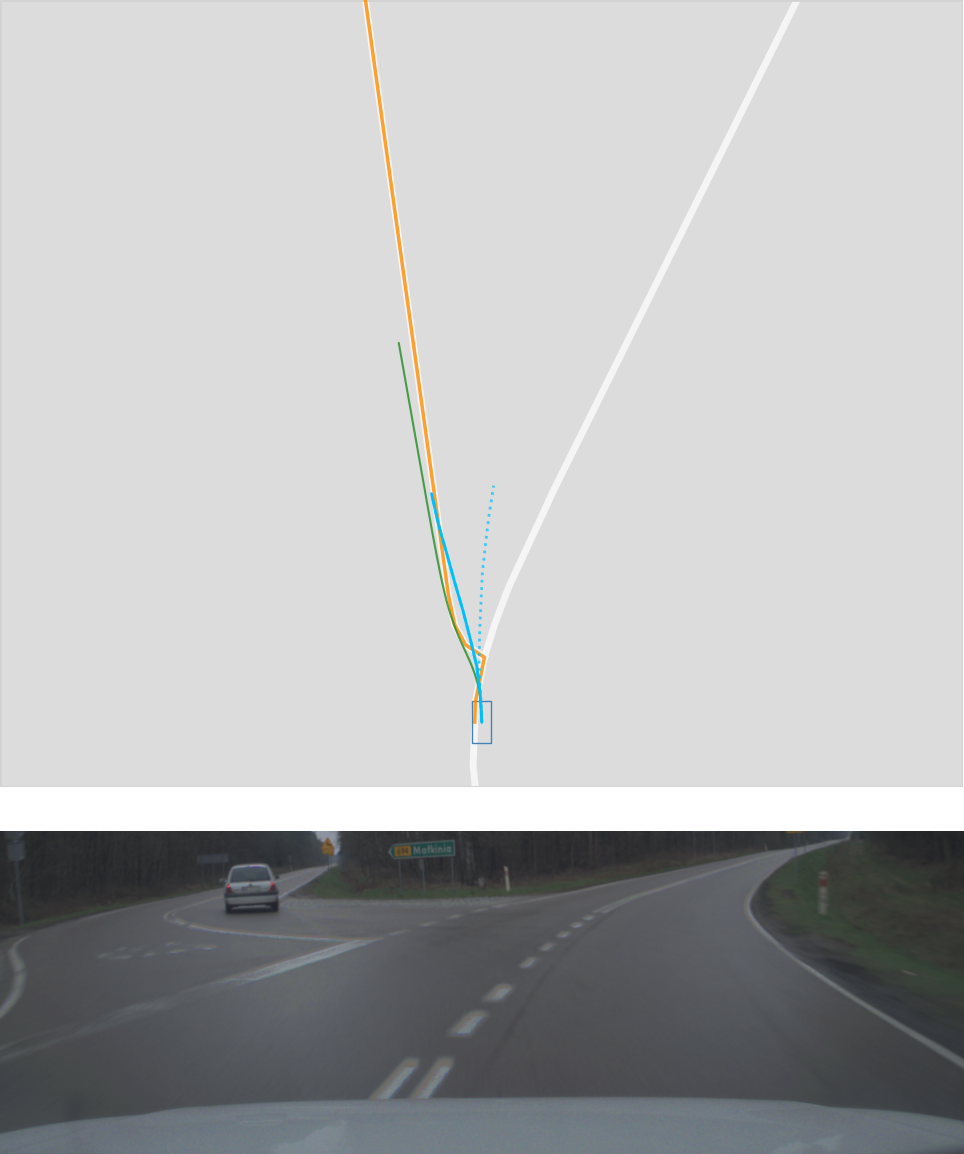}
    \caption{Gating correctly resolves ambiguity in the image, prioritizing route-based prediction.}
    \label{fig:partials_good_2}
\end{subfigure}\hfill
\begin{subfigure}[b]{0.32\textwidth}
    \includegraphics[width=\linewidth]{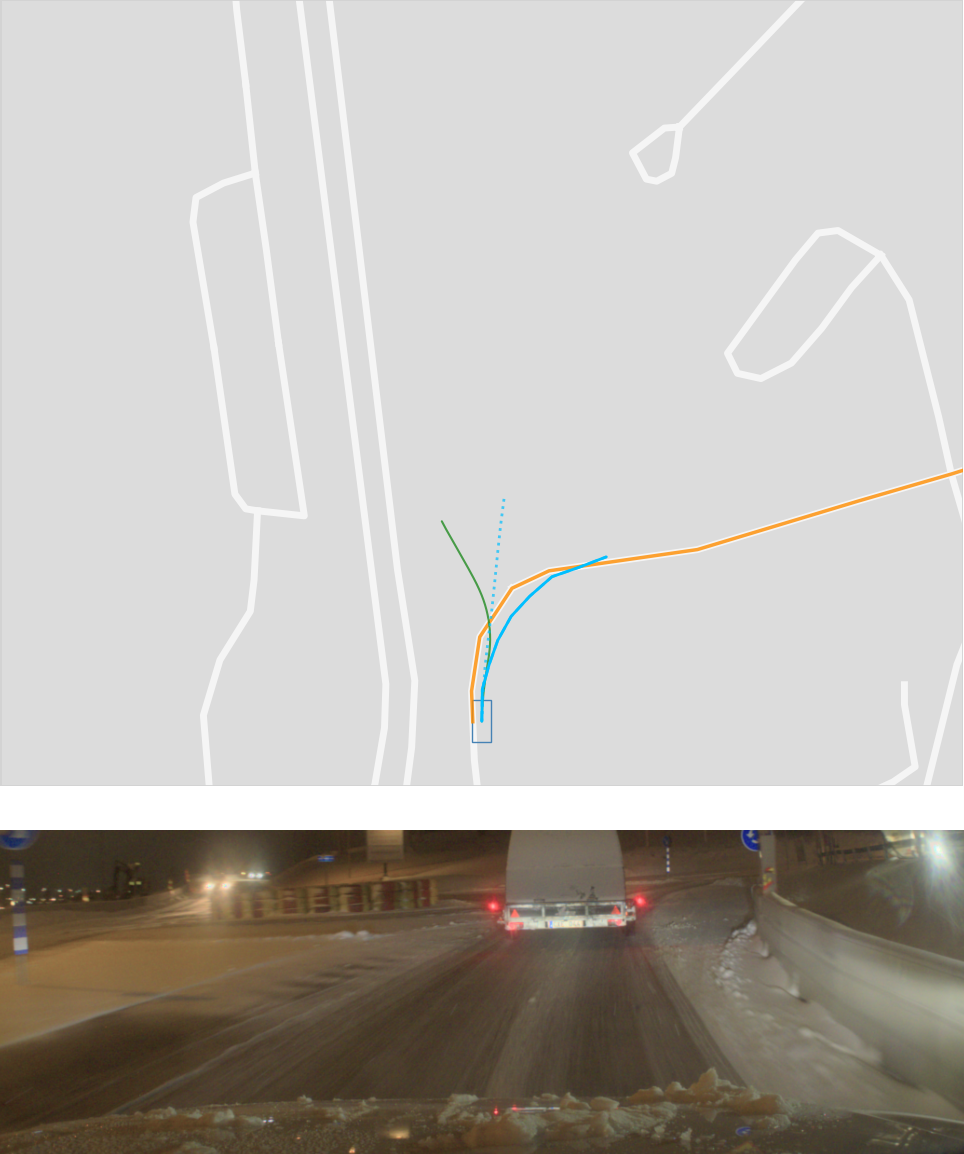}
    \caption{Failure case. Leading vehicle occludes the left bend, while the  route is erroneous.}
    \label{fig:partials_failure}
\end{subfigure}
\caption{Visual comparison of \textcolor{cyan}{\methodname} output, \textcolor{ForestGreen}{ground truth} and the \textcolor{orange}{input route}. Dashed line represents the image-based prediction and in all three cases, the gating picked the route-based output (solid line).}
\label{fig:partial_predictions}
\end{figure*}

\begin{figure*}[t]
\centering
\begin{subfigure}[b]{0.32\textwidth}
    \includegraphics[width=\linewidth]{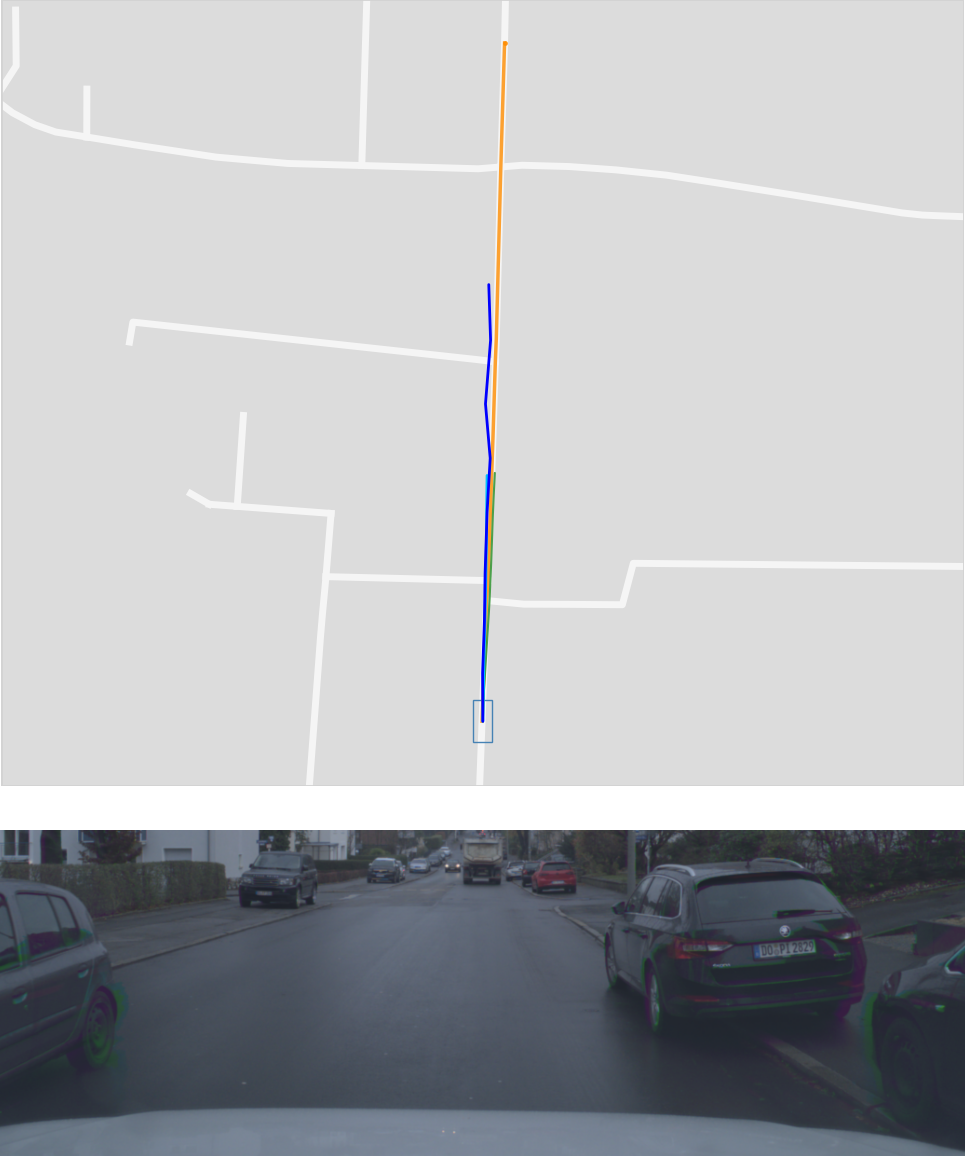}
    \caption{\methodname better accounts for vehicle
    kinematic variation in urban scenarios.}
    \label{fig:models_kinematics}
\end{subfigure}\hfill
\begin{subfigure}[b]{0.32\textwidth}
    \includegraphics[width=\linewidth]{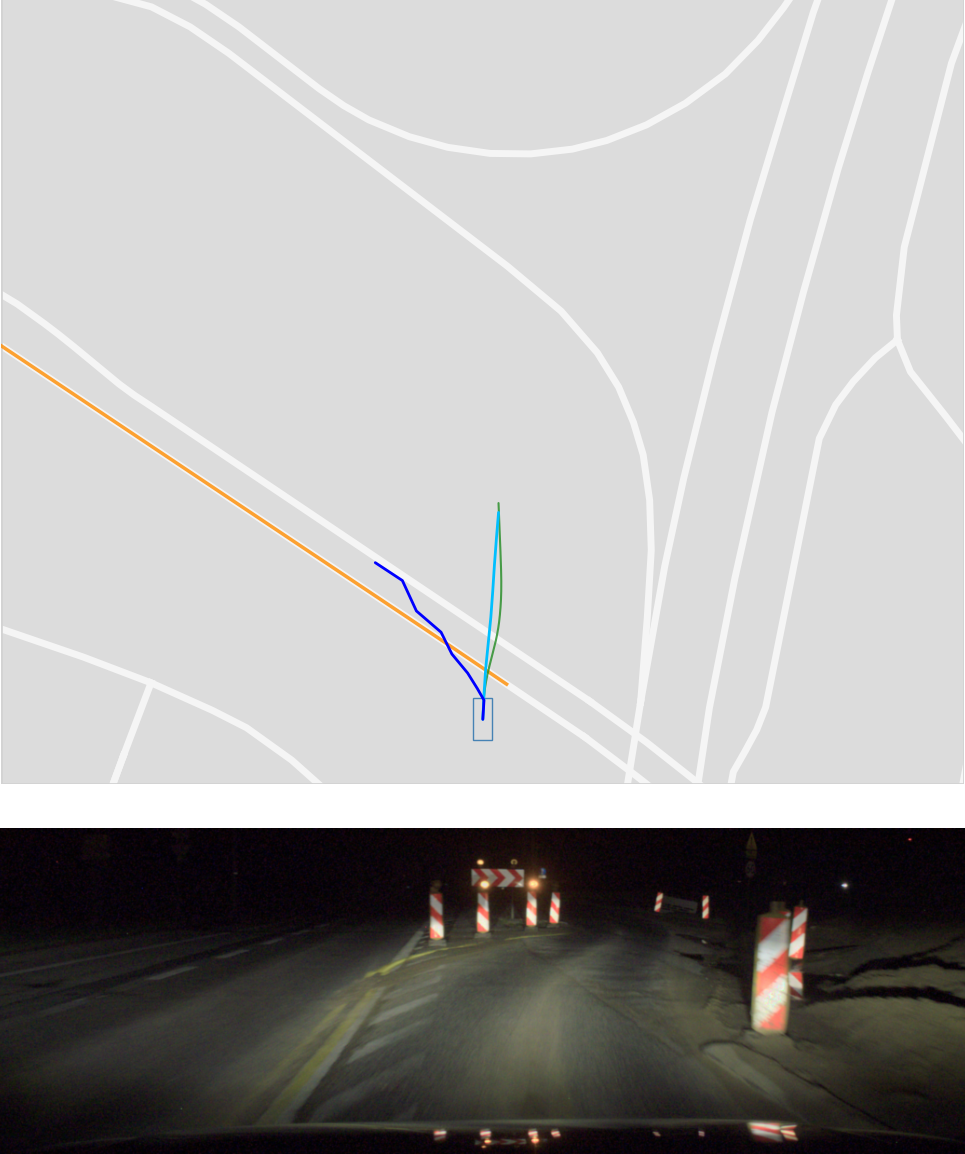}
    \caption{\methodname shows robustness to a mislocalized scene and thus incorrect route.}
    \label{fig:models_loc}
\end{subfigure}\hfill
\begin{subfigure}[b]{0.32\textwidth}
    \includegraphics[width=\linewidth]{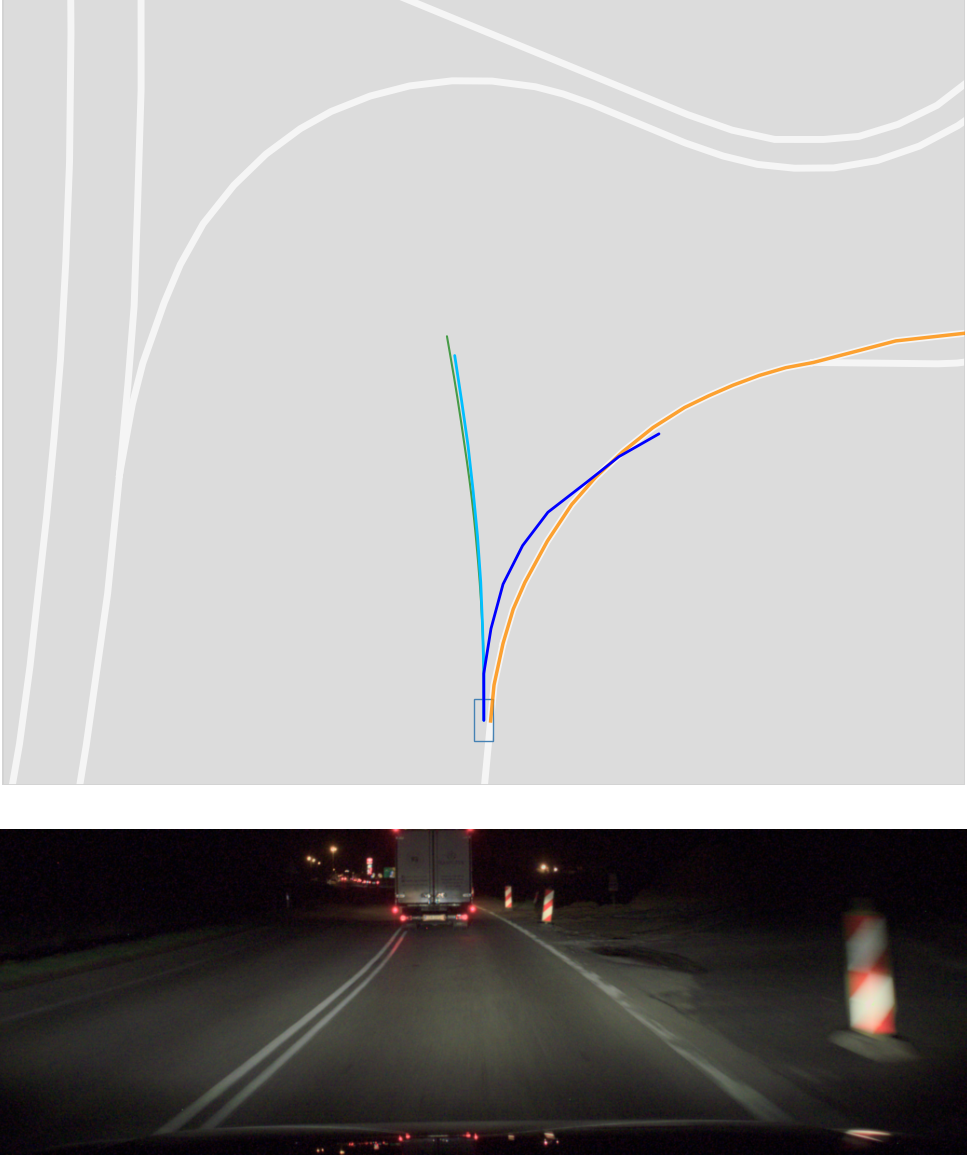}
    \caption{\methodname is robust to a mismatch between SD map and reality (roadworks).}
    \label{fig:models_sdmap}
\end{subfigure}
\caption{Performance comparison of \textcolor{cyan}{\methodname} and \textcolor{navyblue}{I+K+R baseline} in challenging scenarios. \textcolor{ForestGreen}{Ground truth} and \textcolor{orange}{input route} are also visualised.}
\label{modelscomp}
\end{figure*}

\autoref{fig:partials_good_1} and \autoref{fig:partials_good_2} illustrate representative cases where the gating module successfully selected the more reliable modality in the presence of image-level ambiguity. In both examples, the image-based prediction appears plausible in isolation; however, with route information, the model correctly prioritizes the route-based trajectory, which aligns better with the true trajectory. Overall, the gating module selects the route branch in 66\% of test scenarios. \autoref{fig:partials_failure} shows a representative failure case where neither modality provides sufficient information to recover the ground truth. The image is occluded by a leading vehicle, and the leftward curvature of the true trajectory is not visible. Simultaneously, the route input is misaligned due to a localization error, placing the vehicle on an adjacent road. Importantly, such failures require both modalities to be simultaneously misleading, suggesting that the system is generally robust and does not rely on a single point of failure.

\autoref{modelscomp} presents a qualitative comparison between \methodname and our strongest baseline (I+K+R) across three challenging scenarios. \autoref{fig:models_loc} and \autoref{fig:models_sdmap} suggest that \methodname is robust to erroneous route inputs caused by localization errors or by discrepancies between the SD map and the observed scene (e.g., due to infrastructural changes). In contrast, the I+K+R model appears less confident under such misalignments, likely due to its inability to differentially weight input modalities, resulting in less accurate predictions. Furthermore, in urban scenarios shown in \autoref{fig:models_kinematics}, \methodname exhibits improved scene compliance by more effectively leveraging camera-based cues. This is facilitated by its ability to rely more heavily on a single modality when appropriate, rather than being constrained to fuse all inputs early in the architecture regardless of their added value.

\section{Limitations}
One limitation of our study is the imperfect ground‑truth of generated SD‑map routes. These routes can occasionally be inaccurate due to ego‑vehicle localization errors or outdated SD maps, as illustrated in \autoref{fig:models_loc} and \autoref{fig:models_sdmap}. Although more reliable route signals would likely improve performance, \methodname is designed for real‑world deployment, where noisy or incorrect routes are inevitable. We therefore retain such cases in the dataset to evaluate robustness under realistic conditions.

A second limitation concerns data availability. While our ZOD \cite{alibeigi2023zenseact} is publicly released, our experiments rely on an extended internal fleet dataset comprising {\totalsamples} scenarios across 10 European countries and the U.S., which cannot be fully open‑sourced. To support transparency, we release our complete implementation and SD‑map route generation toolkit, enabling reproduction of the full pipeline on the ZOD\cite{alibeigi2023zenseact}. 

%We encourage future work to benchmark SD‑route conditioning on additional public datasets. 

A final limitation is that direct comparison to recent ego trajectory prediction is constrained by dataset and supervision mismatch. State‑of‑the‑art trajectory prediction models typically assume access to HD maps or tracked agents, while our method intentionally does not rely on them. To account for this mismatch, we focus on controlled ablations within a consistent sensor configuration.

\section{Conclusion}
This paper presented \methodname, a lightweight and deployable end-to-end ego-trajectory prediction framework that integrates front-view camera observations, vehicle kinematics, and navigation-route information derived from SD maps. By leveraging a dual-branch architecture and a gating classifier, the proposed method effectively balances short-horizon visual reasoning with long-horizon route intent, while maintaining robustness under route corruption and visual uncertainty.

We evaluated \methodname on a large-scale real-world dataset comprising 480k driving scenarios collected across 10 European countries and the U.S. East Coast, based on an internal extension of ZOD~\cite{alibeigi2023zenseact}. Experimental results demonstrate that our approach achieves a 16.9\% reduction in ADE compared to a strong image-and-kinematics baseline, and exhibits improved robustness in challenging scenarios involving visual occlusions.
%
% Our analysis highlights the complementary roles of the two modalities: front-view images provide rich semantic context for near-field motion understanding, while SD-map route information serves as an effective semantic prior for long-horizon trajectory prediction. These findings suggest that route-conditioned prediction using a widely available SD map is a practical and scalable alternative to HD-map-dependent approaches. 
% As future work, we plan to incorporate surround-view imagery to better model interactions with other road agents and further enhance prediction robustness in complex traffic scenes.

Our analysis highlights the complementary roles of the two modalities: front-view images provide rich semantic context for near-field motion understanding, while SD-map route information serves as an effective semantic prior for long-horizon trajectory prediction. These findings suggest that route-conditioned prediction using widely available SD maps is a practical and scalable alternative to HD-map-dependent approaches.

Several directions remain open for future investigation, including long-term deployment behavior under localization drift, stale map information, navigation inconsistencies, and varying SD-map quality across regions. Another promising direction is tighter integration with downstream planning systems and uncertainty-aware decision making, where confidence in route priors and visual observations could adapt prediction behavior online.

%This paper presented \methodname, a lightweight and deployable end-to-end trajectory prediction framework that leverages a front-view camera, vehicle kinematics, and route information extracted from an SD map. The architecture employs a gating classifier to dynamically select the most suitable trajectory prediction from two specialized branches: either image-based or route-based. We conducted extensive evaluations on an internal extension of ZOD \cite{alibeigi2023zenseact} comprising 480,000 driving scenarios collected across 10 European countries and the U.S. East Coast. Experimental results demonstrate that our proposed method achieves a 16.9\% improvement in Average Displacement Error (ADE) over the baseline method. Furthermore, the model exhibits robustness in scenarios with visual occlusions.
%Our analysis reveals that front-view images provide rich semantic context in the near-field around the ego-vehicle, while route information effectively supports long-horizon predictions. As future work, we plan to incorporate surround-view images to better account for other road agents.

\bibliographystyle{IEEEtran}
\bibliography{references}  % .bib
\end{document}